\documentclass{article}
\usepackage{ijcai22}

\usepackage{times}
\usepackage{soul}
\usepackage{url}
\usepackage[hidelinks]{hyperref}
\usepackage[utf8]{inputenc}
\usepackage[small]{caption}
\usepackage{graphicx}
\usepackage{amsmath}
\usepackage{amsthm}
\usepackage{booktabs}
\usepackage{algorithm}
\usepackage{algorithmic}
\usepackage[T1]{fontenc} 
\usepackage{marvosym} 
\title{Quantum Graph Learning: Frontiers and Outlook}

\author{
Shuo Yu$^1$
\and
Ciyuan Peng$^{2,3}$
\and
Yingbo Wang$^4$
\and
Ahsan Shehzad$^4$
\and
Feng Xia$^5\textsuperscript{(\Letter)}$
\and
Edwin R. Hancock$^6$
\affiliations
$^1$School of Computer Science and Technology, Dalian University of Technology, China\\
$^2$School of Information Management and Business Administration, Chengdu Neusoft University, China\\
$^3$Institute of Innovation, Science and Sustainability, Federation University Australia, Australia\\
$^4$School of Software, Dalian University of Technology, China \\
$^5$School of Computing Technologies, RMIT University, Australia\\
$^6$Department of Computer Science, University of York, UK\\
\emails
shuo.yu@ieee.org,
ciyuan.p@outlook.com, 
zqqgpllp20@gmail.com, 
ahsan.shehzad@outlook.com, 
f.xia@ieee.org, 
edwin.hancock@york.ac.uk}

\begin{document}

\maketitle

\begin{abstract}
Quantum theory has shown its superiority in enhancing machine learning. However, facilitating quantum theory to enhance graph learning is in its infancy. This survey investigates the current advances in quantum graph learning (QGL) from three perspectives, i.e., underlying theories, methods, and prospects. We first look at QGL and discuss the mutualism of quantum theory and graph learning, the specificity of graph-structured data, and the bottleneck of graph learning, respectively. A new taxonomy of QGL is presented, i.e., quantum computing on graphs, quantum graph representation, and quantum circuits for graph neural networks. Pitfall traps are then highlighted and explained. This survey aims to provide a brief but insightful introduction to this emerging field, along with a detailed discussion of frontiers and outlook yet to be investigated.\end{abstract}

\begin{figure}[!h]
	\includegraphics[scale=0.37]{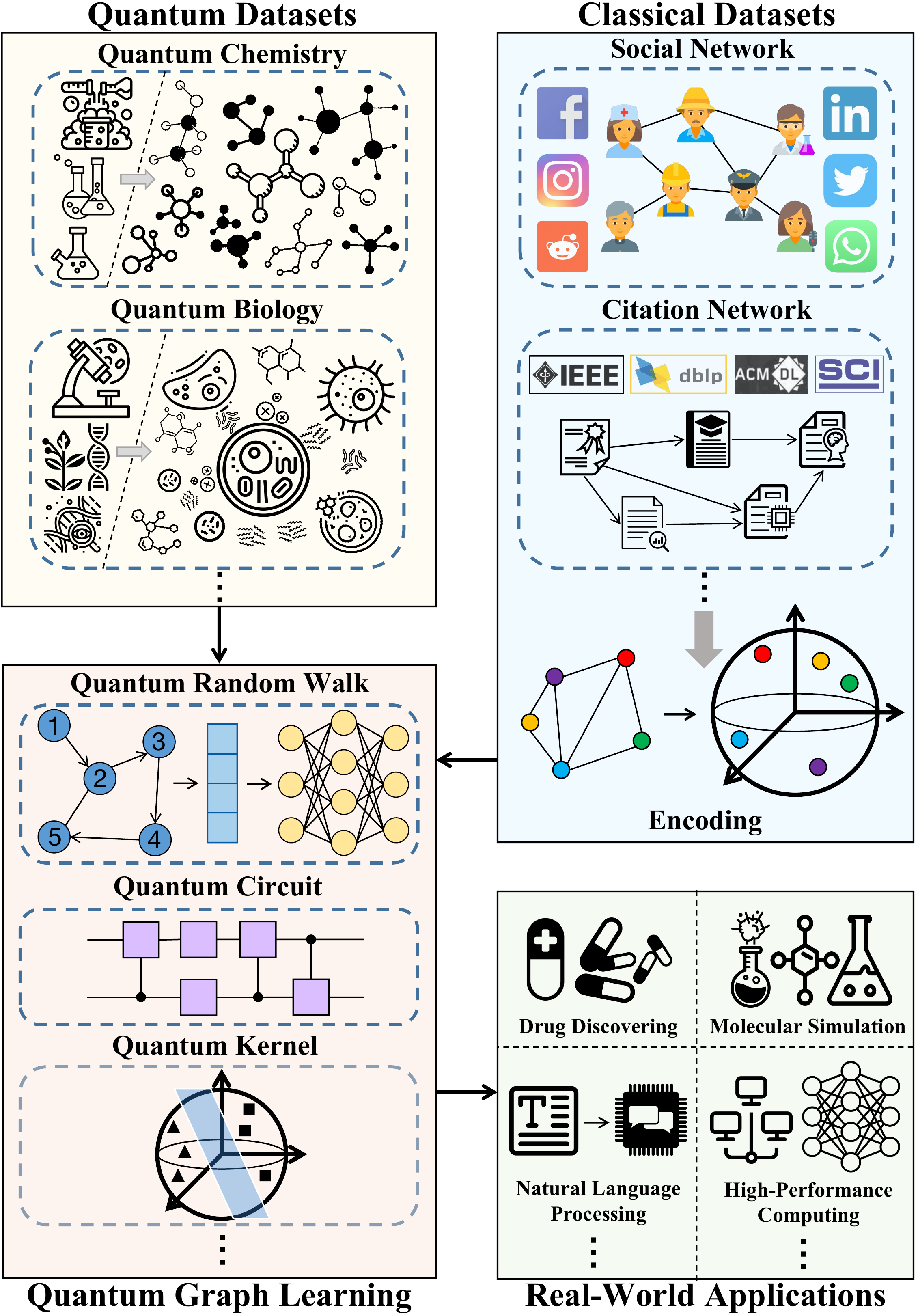}
	\centering
	\caption{An illustration of how general quantum graph learning works.}
	\label{fig:2}
\end{figure}

\section{A Look at Quantum Graph Learning}

The increasing number of new studies in quantum theory expands the boundary of both physics and computer science. From rethinking fundamental concepts to conceiving revolutionary new technologies, quantum theory has been investigated with tremendous progress, which gives further directions to many scientific research areas, including graph learning. The quantum potential has been widely verified in machine learning~\cite{dunjko2018machine,liu2021rigorous}. The advent of hardware and the high-performance simulation framework of quantum has prompted the research progress of graph learning~\cite{beer2021quantum}. Quantum graph learning (QGL) exploits quantum computational power and quantum theoretical basis to provide solutions for graph learning. The concept of QGL shares overlapping fields with physics-inform graph learning (PIGL) ~\cite{karniadakis2021physics}. The QGL offers a new way to encode graph data with qubits and embeds them onto quantum states, whereas other physics theories cannot.

Figure~\ref{fig:2} briefly introduces how general quantum graph learning currently works. Unlike graph learning, QGL can also directly handle quantum datasets (e.g., quantum chemistry and biology datasets) as input. Therefore, for QGL, two different input datasets are feasible. By quantum mechanics, such as the well-known quantum random walk, QGL converts input data to quantum states and represents them with the full function of quantum mechanics’ superpower, i.e., quantum state superposition, quantum entanglement, etc. As a result, QGL has shown its power and potential in representing graph-structured data compared to traditional graph learning. Therefore, QGL will thus be benefiting real-world applications correlated to graphs, including natural language processing, drug discovery, high-performance computing, etc.

As an emerging topic, QGL has drawn rising attention. While current studies are in their initial stage, the blossom of QGL needs far more attention. Some survey papers focus on Quantum Machine Learning (QML), which illustrates both the advantages and challenges of employing quantum mechanics in machine learning~\cite{cerezo2022challenges}. It is recognized that QML will revolute machine learning with quantum hardware and quantum computing. Coincidentally, some survey papers focus on quantum computing on graphs, which introduces how quantum theories accelerate the computing efficiency on graphs~\cite{tang2022quantum}. However, there still needs to be a comprehensive survey paper of QGL that introduces the underlying reciprocity of quantum theory and graph learning and meanwhile points out the pitfall traps and future research directions for both beginning and junior scholars. To fill the blank, we investigate current studies on QGL and give out this survey. This survey shed light on QGL from the perspectives of concept, essence, method, pitfall, and future direction.\\
\textbf{Contributions.} (1) We reveal the mutualism of quantum theory and graph learning. Instead of directly introducing QGL, we first discuss why quantum theory and graph learning naturally can be an effective combination from the perspectives of the characteristics of graph-structured data, the bottleneck of graph learning, and the reciprocity. This can be an easy start for those who are interested. (2) We propose a taxonomy of quantum graph learning. We categorized previous studies on QGL into three kinds: quantum computing on graphs, quantum graph representation, and quantum circuit for graph neural networks.
(3) We point out the pitfall traps for beginners. Three pitfall traps are given in our survey, respectively. We should clarify that QGL is not difficult to get started. Besides, QGL is not a new gimmick of QML. Last but importantly, quantum is not omnipotent for graph learning.

\section{The Mutualism of Quantum Theory and Graph Learning}
\label{sec:mutualism}

Quantum theory has its natural advantages in handling graph data. The superiority of quantum theory in high-performance computing factually renews the way of dealing with graph-structured data, thus providing a new paradigm for graph learning. Looking from the other side, the rigorous requirements of storing, accessing, and processing graph learning tasks conform to the ample scope of quantum theory's abilities. The marriage of quantum theory and graph learning fully exploits the role of quantum theory and tackles the bottleneck of graph learning. The characteristics of graph data, the bottleneck of graph learning, and an illustration of the reciprocity of quantum theory and graph learning are presented.

\subsection{The Characteristics of Graph-structured Data}

The computing complexity of graph computing motivates the development of graph learning to a great extent. One of the underlying reasons is the limitation of computing resources to handle graph-structured data ~\cite{10.1145/3487553.3524718,9416834}. There are three traditional data structures to represent graph-structured data, including an adjacency list, adjacency matrix, and incidence matrix. An adjacency list uses a linked table as a basic data structure to represent graph data. An adjacency matrix is a two-dimension matrix wherein each element represents the existence of an edge between two specific nodes. An incidence matrix is a typical logic matrix representing relations between each node and every other edge. However, the requirement of storing and using graph-structured data is far more rigorous, no matter in academia or industry~\cite{9052480}. Even though some industry-friendly (e.g., dense\_hash\_map in SparseHash developed by Google) or academic-friendly (e.g., sparsematrix in PyTorch) graph structures have been developed, handling graph data is still a challenging task compared to other data structures, not mention to dynamic graphs~\cite{DBLP:journals/tnse/XiaYLLL22}.

The requirements for handling graph data are two-fold, i.e., space and speed. A smaller footprint is needed to hold more significant amounts of graph data. At the same time, there is also a need for low access latency and friend concurrent access. Such a dilemma notably prompts the development of graph embedding and graph representation learning, which, unfortunately, is a trade-off solution for handling graph data. 

\subsection{The Bottleneck of Graph Learning}

Graph learning has made significant progress, but it also comes to its bottleneck. Graph learning cannot aggregate information over long paths, which causes exponentially growing data to be compressed into fixed-size vectors~\cite{yu2022graph}. Consequently, when the downstream task relies on remote information interactions, information from more distant nodes cannot be propagated. Though some graph learning methods can integrate global information, they are compensatory approaches for the lack of learning ability in graph learning~\cite{DBLP:conf/cikm/CaoLX15}. In other words, the need for learning ability in graph learning methods still exists. Another bottleneck of graph learning is explainability and interpretability. They are two different concepts. An interpretable (i.e., white box) graph learning model refers to that it can provide a human-understandable explanation for its tasks~\cite{DBLP:conf/icml/MiaoLL22}. While for a black box model, if the output results can be understood by post hoc explanation techniques, then it is explainable. With the help of quantum theory, no matter whether the explainable or interpretable problem will be resolved in graph learning. In addition, graph learning has limited ability to propose practical solutions for large-scale graphs (e.g., brain network). An eclectic solution is first partitioning the large-scale graph into many smaller pieces and then unifying the representation learning or implementing parallel methods of each component.

\subsection{The Reciprocity of Quantum Theory and Graph Learning}

Quantum theory provides the theoretical foundation of quantum computing, a new computational paradigm. By exponentially accelerating the traditional computing methodologies, quantum computing shows its promising prospective and potential superiority. Quantum computers have the natural advantage in integer factorization and discrete logarithm~\cite{DBLP:conf/focs/Shor94}. Moreover, especially in unstructured databases, quantum computing also outperforms traditional computing methods in searching and retrieving target values~\cite{DBLP:conf/stoc/Grover96}. All these advantages give impetus to merging quantum computing and graph learning.

The nonlocal effect in quantum theory refines how we think and perceive the world. Quantum teleportation exploits the nonlocal effect to transmit a quantum state across extremely long distances with negligible communication cost~\cite{PhysRevLett.123.070505}. Similarly, global information can be transferred with very short path (e.g., six degrees of separation in social graphs) in graph-structured data~\cite{DBLP:journals/csr/YuFZBXX20,DBLP:journals/access/YuXZXAT19}. Moreover, because of the non-Euclidean data structure, global information of graphs also occupies a critical position in graph learning~\cite{DBLP:conf/jcdl/XuYSRLP020}. But current message passing techniques of graph learning cannot integrate long-distance or global information precisely with low computational costs (e.g., time costs). 

QGL can represent the input data in the form of quantum states, therefore, the message passing mechanism of QGL involves coherence or entanglement, which can help transfer long-distance information with a faster time. Moreover, graph learning enhanced with quantum theory has shown promising potential in solving the large-scale problem. With quantum-friendly hardware, qubit encoding can lead to greater information capacity with lower calculation requirements. In addition, the criticism of the black box graph learning model will be resolved by employing quantum theory. Quantum theory provides a theoretical basis for large-scale graph learning and profoundly impacts real-world applications (e.g., cognitive function). As a result, whole graph computing or representation learning will become a reality and improve downstream tasks' accuracy. The completeness of quantum theory can fully support the computing, representation, and learning of graph data.

The transparency and interpretability of graph learning models will be ensured with the introduction of quantum theory, thus benefitting graph data storage, access, and processing. Though quantum computing for graph learning is still in its early stage, the huge potential of tackling graph problems has shown. The mutualism of quantum theory and graph learning is stand to reason.

\section{A Taxonomy of Quantum Graph Learning}

In this survey, we discuss quantum graph learning methods in three categories: quantum computing on graphs, quantum graph representation, and quantum circuits for graph neural networks. Table~\ref{category} lists the representative methods of these three categories.

\begin{table*}
	\centering
	\begin{tabular}{p{0.75\columnwidth}lllp{0.4\columnwidth}}
		\toprule
		Type & Method&Input&Application\\
		\midrule
		Quantum Computing on Graphs 
		& SEQO~\cite{tabi2020quantum}& Q  &Graph Coloring\\
		& QAD~\cite{pelofske2021decomposition}& Q&Maximum Clique; Vertex Cover\\
		&QAISV~\cite{wang2022quantum}& Q &Graph Partitioning\\
		
		Quantum Graph Representation 
		&QS-CNN~\cite{zhang2019quantum}&C  &Node Classification \\
		&QGRE~\cite{yan2022towards}& Q&Node Classification\\
		& QSGCNN~\cite{bai2021learning}  & C  &Graph Classification \\
		&  QEK~\cite{henry2021quantum}  & C   &Graph Classification\\
		&  QSGK~\cite{kishi2022graph}  & C   &Graph Classification \\
	Quantum Circuit for Graph Neural Networks  	&HQGNN~\cite{tuysuz2021hybrid}&C&Link Prediction\\
	& DQGNN~\cite{ai2022decompositional}& Q &Graph Classification\\
		&  QGCL~\cite{chen2022novel} & Syn.    &Node Classification\\
		& EQGCs~\cite{mernyei2022equivariant}& Q   &Graph Classification\\

		\bottomrule
	\end{tabular}
	\caption{Quantum graph learning methods. ‘Q’ is quantum, ‘C’ is classical, and ‘Syn.’ is synthetic. The input data is ‘Q’, ‘C’, and ‘Syn.’ respectively meaning that the corresponding data processing is based on the classical computer, quantum computer, and quantum computer assisted by classical computing modules.}
	\label{category}
\end{table*}

\begin{figure}[h]
	\includegraphics[scale=0.37]{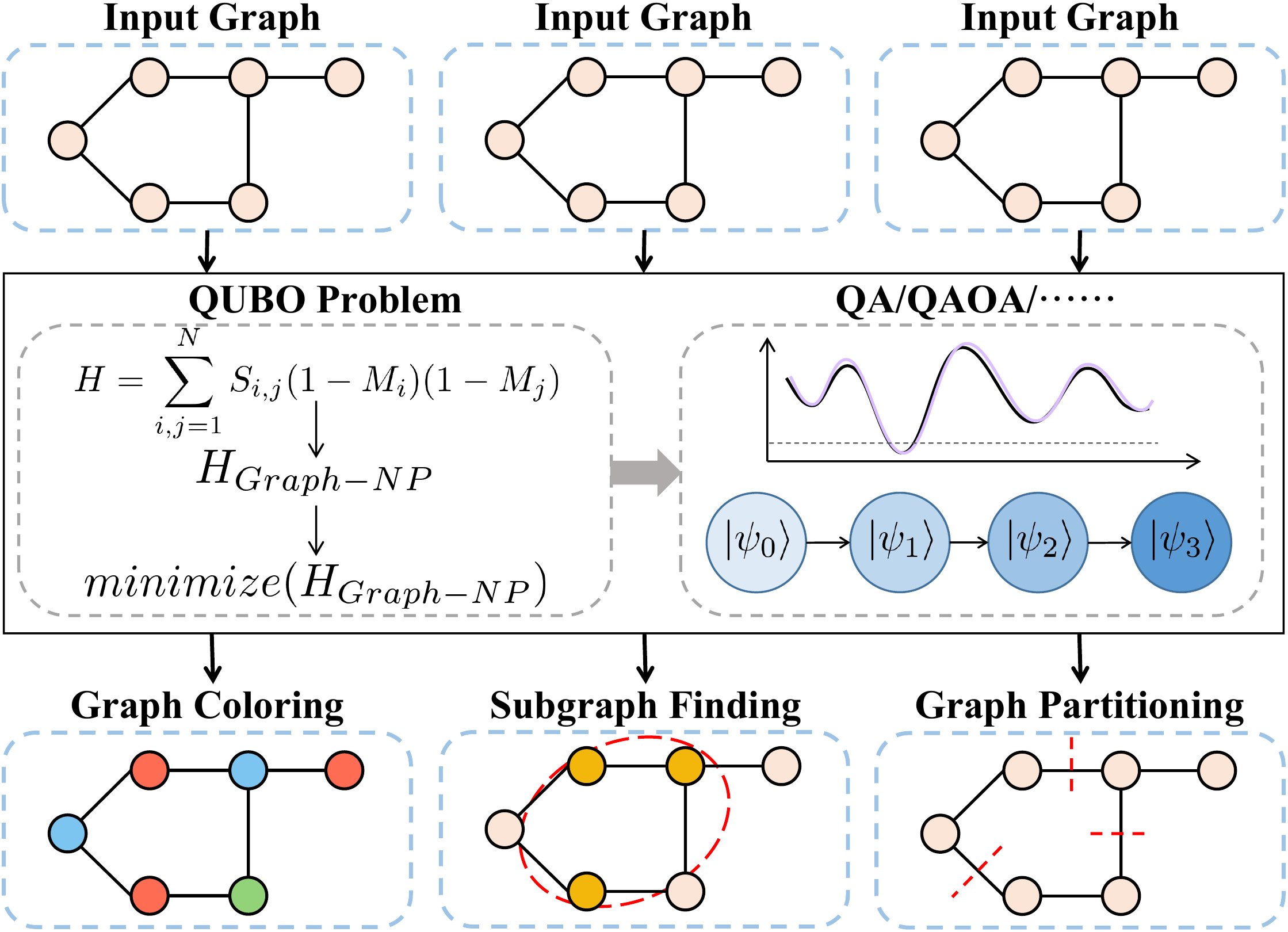}
	\centering
	\caption{Quantum Computing on Graphs.}
	\label{fig:3}
\end{figure}
\subsection{Quantum Computing on Graphs}
 
Implementing quantum computing on graph data is one of the standard approaches for data preprocessing and graph theoretical problems, especially NP-hard graph problems, such as graph coloring, vertex cover, and graph partitioning. 
The typical method is performing quantum evolution on the underlying graph structure to describe the topology information of graphs~\cite{tang2022quantum}. 
The formulations of many graph-related NP-hard problems are Quadratic Unconstrained Binary Optimization (QUBO) problems, which can be as the input for Quantum Annealing (QA) and Quantum Approximate Optimization Algorithm (QAOA)~\cite{lucas2014ising}. Figure~\ref{fig:3} shows the example of quantum computing-based algorithms for NP-hard graph problems.~\cite{tabi2020quantum}. The form of QUBO problems is the minimization of $f$:

 \begin{equation}
 	\begin{aligned} 		
 	\label{QUBO}
 	\mathop{minimize}f(x)=\sum_{i,j=1}^N S_{i,j}x_ix_j,\\
 	y=\mathop{\arg\min}\limits_{x \in \{0,1\}^N}f(x).
 	 \end{aligned}
 \end{equation}
 Here, $f(\cdot)$ is the cost function and $S$ indicates a real symmetric matrix. $y$ denotes a global minimizer of $f(\cdot)$.

The QUBO problems described above can be regarded as finding the minimum energy of $N$-qubit Ising Hamiltonian, which describes the dynamics of quantum systems. The definition of Hamiltonian is as follows:
 \begin{equation}
		\label{Hamiltonians}
		H=\sum_{i,j=1}^{N}S_{i,j}(1-M_i)(1-M_j),
\end{equation}
where $M_z$ is the operator acting as the Pauli-$M$ gate on the $z$th qubit.

Many studies have contributed to NP-hard graph problems by formulating them into QUBO problems and utilizing quantum computing to obtain the solutions. For the graph coloring problem, Tabi et al. proposed a space-efficient quantum optimization algorithm (SEQO) based on quantum annealing~\cite{tabi2020quantum}. They considered the graph coloring problem a QUBO problem, then applied QUBO as an input for QAOA. They designed a flexible approach to reduce the number of qubits to provide a space-efficient solution for graph coloring. More recently, Pelofske et al. designed a QA-based decomposition algorithm (QAD) for the maximum clique and the minimum vertex cover problems~\cite{pelofske2021decomposition}. Particularly, they aimed to obtain an optimal set of vertice by minimizing Ising Hamiltonian. Their model recursively splits the given instance into small subproblems, which can be solved directly by utilizing a quantum annealer. Wang et al. focused on the partitioning problem of the grid, which is a kind of graph ~\cite{wang2022quantum}. To handle inequality constraints in the grid partitioning optimization model, they proposed a framework based on quantum annealing with integer slack variables (QAISV). They especially implemented the integer slack and binary expansion methods to transform the grid partitioning problem into a QUBO problem. Moreover, other NP-hard graph problems, such as graph isomorphism~\cite{tang2022quantum}, can also be reliably solved by applying quantum evolution via QUBO.

\subsection{Quantum Graph Representation}

\begin{figure*}[!h]
	\includegraphics[scale=0.32]{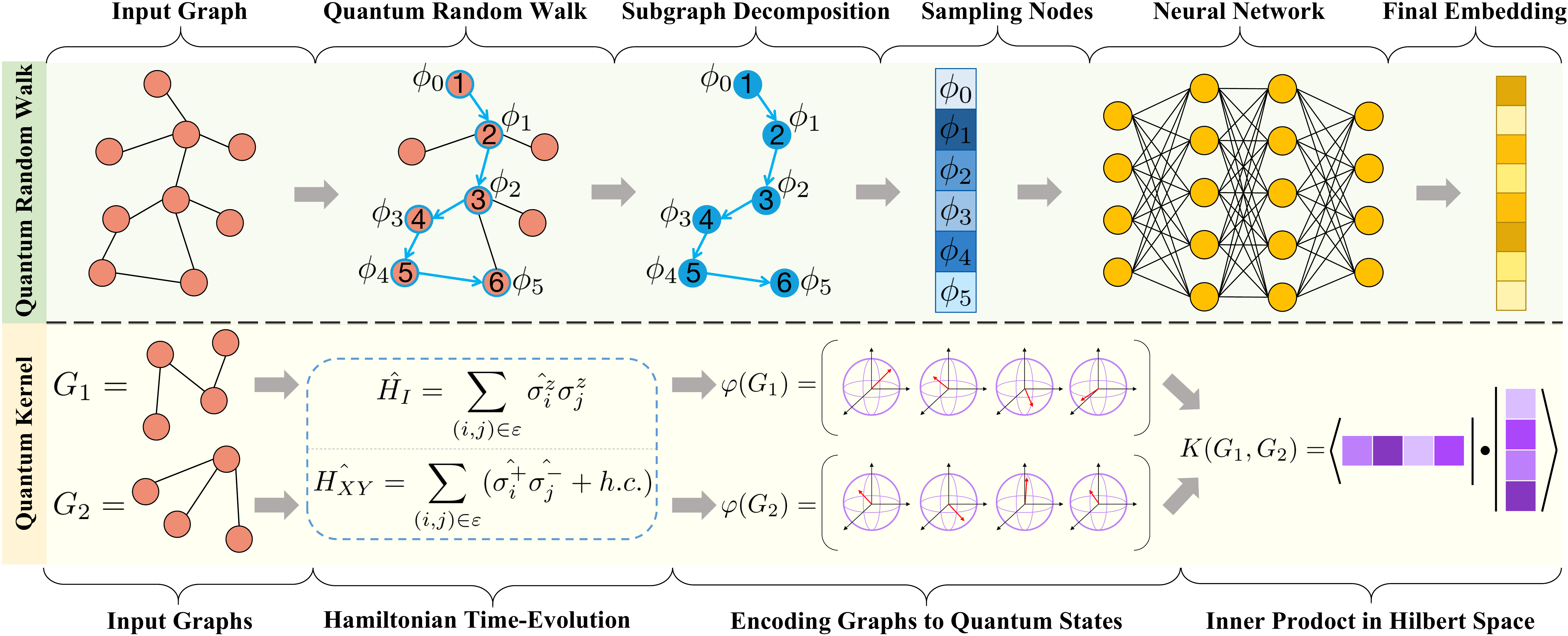}
	\centering
	\caption{Quantum graph representation learning with quantum random walks and quantum kernels.}
	\label{fig:4}
\end{figure*}

The rapid development of quantum computing has recently opened up new opportunities for graph representation. Graph representation, mapping graphs into embedding vector space for facilitating various downstream tasks (e.g., link prediction and combinatorial optimization), has attracted extensive attention. Many quantum graph representation algorithms have proved their significant capabilities in extracting unseen or atypical patterns in graphs~\cite{huang2021power}. Specifically, the core idea of quantum graph representation is embedding graphs to quantum states, of which the entanglement and superposition can effectively characterize the features of graphs in a Hilbert space~\cite{cong2019quantum,schuld2019quantum}. Two representative approaches are quantum random walks and quantum graph kernels. Figure~\ref{fig:4} gives examples of them. In the rest of this section, we will discuss these two methods in detail.

\subsubsection{Quantum Random Walks}

In quantum random walks, the initial distribution of the walker is characterized by the amplitudes of quantum states and then the walker can evolve in a quantum mechanical manner. Given a graph $G(V,E)$, where $V$ and $E$ are vertex set and edge set respectively, in the Hilbert space $ \mathcal{H}$, the basic state of the walker at vertex $v\in V$ is $\boldsymbol{\eta_v}$. The quantum state of the walker $|\psi(t) \rangle$ at time $t$ can be indicated as a linear combination of $\boldsymbol{\eta}$:

\begin{equation}
	\label{quantum sate of walker}
	|\psi(t) \rangle=\sum_{v\in V}\alpha_v\boldsymbol{\eta_v},
\end{equation}
wherein, $\alpha_v$ is the complex amplitude. Particularly, there are two types of quantum random walks: discrete time and continuous time quantum random walks~\cite{tang2022quantum}. In discrete time quantum random walks (also called coin quantum random walks), two Hilbert spaces need to be defined. One is the position Hilbert space $ \mathcal{H}_p$, which captures the superposition of nodes. Another one is the coin Hilbert space $ \mathcal{H}_c$ capturing the multi-direction superposition of the walker on each node. Generally, the space of the quantum random walks can be described as $\mathcal{H}=\mathcal{H}_p\bigotimes\mathcal{H}_c$. Different from discrete time quantum random walks, continuous time quantum random walks only consider the position Hilbert space $ \mathcal{H}_p$.

Various works adopting quantum random walks for graph representation have been proposed. For example, Zhang et al. proposed a quantum-based subgraph convolutional neural network (QS-CNN)~\cite{zhang2019quantum}, which introduces a graph decomposition method based on quantum random walks. In particular, quantum random walks capture different patterns of vertices connectivity by destructive and constructive interference. Then the graph is decomposed into a family of multi-layer expansion subgraphs in the vector space.

Yan et al. presented a method, namely Quantum Graph Recurrent Embedding (QGRE), which aims to adopt quantum random walks to attributed graphs~\cite{yan2022towards}. Their model first applied discrete-time quantum random walks to a node-attributed graph, and then the quantum state sequences of the graph are fed into a quantum LSTM to obtain node embeddings. On the contrary, Bai et al. took advantage of continuous-time quantum random walks to capture graph characteristics~\cite{bai2021learning}. They proposed a quantum spatial graph convolutional neural network (QSGCNN), which can use quantum vertex information propagation to extract multi-scale node features.

\subsubsection{Quantum Graph Kernels} 

As the Hilbert space of quantum states can be a feature space where the graph kernels are induced, quantum graph kernel-based graph representation methods have emerged in recent years. Quantum graph kernels aim to represent different graphs in a Hilbert space and then compare the similarity between two graphs in terms of the quantum representations~\cite{schuld2020measuring}. Specifically, a quantum graph kernel constitutes an inner product to measure the similarity of graphs. Given two graphs $G_1$ and $G_2$, there is a $\varphi $ mapping two graphs into a Hilbert space $ \mathcal{H}$, and the similarity between $G_1$ and $G_2$ can be measured by a kernel $K$:

\begin{equation}
	\label{kernel}
	K(G_1,G_2)=\langle\varphi(G_1)|\varphi(G_2)\rangle.
\end{equation}

An increasing number of works have exploited the area of quantum graph kernel-based graph representation. A quantum evolution kernel (QEK) has been presented to characterize graphs~\cite{henry2021quantum}. The core idea of QEK is taking Hamiltonian encoding-based quantum evolution as a tool to realize a graph kernel. Kishi et al. proposed a quantum superposition-based graph kernel (QSGK) to extract the features of subgraphs and measure their similarity~\cite{kishi2022graph}. In their method, the main motivation is to efficiently map many subgraphs into a quantum state in the Hilbert space, and then achieve high-performance downstream tasks (e.g., graph classification).

\begin{figure*}[h]
	\includegraphics[scale=0.39]{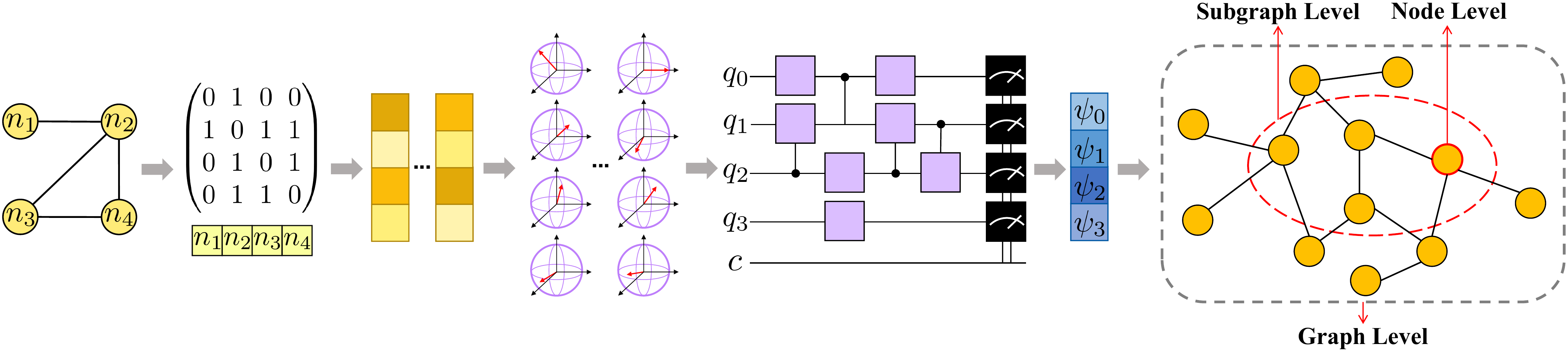}
	\centering
	\caption{Quantum circuits for graph neural networks.}
	\label{fig:5}
	\vspace{-0.3cm}
\end{figure*}

\subsection{Quantum Circuits for Graph Neural Networks}

With the development of noisy intermediate-scale quantum (NISQ) devices, more and more researchers have focused on quantum graph neural networks, which combine the graph neural networks with quantum modules to optimize the current models~\cite{verdon2019quantum}. As a result, quantum graph neural networks have shown various advantages, including reducing the number of training parameters and the complexity of learning models. There are two kinds of approaches to implementing quantum graph neural networks. One is applying fault-tolerant quantum computer-based quantum algorithms to accelerate the calculation step of classical graph neural network models. Another widely concerned method is based on NISQ devices and modifying the structure of graph neural networks with quantum circuits~ \cite{chen2022novel}. The mainstream quantum structure deployed on NISQ devices is parameterized quantum circuits (PQCs), of which the components are multiple adjustable quantum gates (e.g., $R_y(\theta)$ gate) and fixed quantum gates (e.g., Pauli-$Z$ gate)~\cite{araujo2021divide}. The objective function of learning models would be approximated by adjusting the quantum gate circuit parameters. Significantly, in PQCs-modified graph neural networks, the input graph data is encoded into quantum amplitudes. Figure~\ref{fig:5} is the illustration of quantum graph neural networks. 

A hybrid quantum-classical graph neural network (HQGNN) has been proposed for particle track reconstruction problems~\cite{tuysuz2021hybrid}. HQGNN considered two types of PQCs with different entangling capacities and expressibility to obtain desired outputs. One of the PQCs types consists of circuits with hierarchical architectures, while another is composed of parametrized gate layers. More recently, Ai et al. presented a decompositional quantum graph neural network (DQGNN) using a fixed-sized quantum device to handle larger-sized graph data~\cite{ai2022decompositional}. They aimed to alleviate the challenge of the limited number of physical qubits by using the unitary matrix representation and tensor product to reduce the number of required parameters. Particularly, they designed a quantum circuit with a hierarchical structure (containing multiple layers) to capture node information. Chen et al. designed a quantum graph convolutional layer (QGCL) to realize quantum graph convolutional networks~ \cite{chen2022novel}. They presented a novel PQC by implementing a linear combination unitary-based adjacent matrix in QGCL. Then, they utilized a series of quantum gates with adjustable phases to obtain a weight matrix. Mernyei et al. proposed equivariant quantum graph circuits (EQGCs) as a class of PQCs~\cite{mernyei2022equivariant}. In detail, they designed two subclasses of EQGCs, including equivariant Hamiltonian quantum graph circuits (EH-QGCs) and equivariantly diagonalizable unitary quantum graph circuits (EDU-QGCs), to provide a unifying framework for quantum graph neural networks.

\section{Pitfalls}

Quantum theory shows its potential power in enhancing and reforming graph learning. We point out some pitfall traps in this section for those who aim to dig into this research area.

\subsection{It is Difficult to Get Started}
Quantum mechanism is counter-intuitive and perplexing, but only some fundamental parts of quantum computation and quantum information are essential to start QGL. For quantum computation, beginners should fully recognize linear algebra and complex numbers, which are also required for graph learning. The reason is that the essence of quantum states passing through quantum circuits is just the product of complex matrices. As for quantum-related preparation knowledge, quantum computing and quantum teleportation need to be grasped. The understanding of quantum computing will help to build initial recognition of static resources (i.e., qubit), and that of quantum teleportation will be conducive to comprehending the dynamic processes (e.g., transmission and copy), respectively~\cite{nielsen2002quantum}. Specifically, Hilbert space can be regarded as the generalization of finite-dimensional Euclidean space, which is expanded to infinite dimensions and complex numbers. Therefore, such ultra-high dimension does not fit downstream tasks. In quantum-classical hybrid algorithms, quantum circuits output quantum states to classical computers. Ai et al. computed the von Neumann entropy of quantum states in Hilbert space and represented them in low-dimensional Euclidean space~\cite{ai2022decompositional} for graph classification. Thus, quantum information can be exploited to characterize the outputted quantum states for downstream tasks.

\subsection{QGL is a New Gimmick as a Branch of QML}
It is known that graph-structured data is non-Euclidean data, which brings about difficulties and challenges for graph learning. Because of such specificity, graph learning differs from other branches of deep learning. This specificity of graph-structured data thus also makes QGL different from QML. 

Graph learning has rigorous requirements for efficiency in storage and computation, while quantum memory, quantum searching, and quantum computing methodologies can fundamentally fix those problems. Classical hardware limits the ability to access or process sparse graph-structured data efficiently. Consequently, the large scale of graph-structured data has been a longstanding challenge. The computational inefficiency of non-Euclidean graph data becomes the main barrier to the development of graph learning. In contrast, quantum computing provides an exponential acceleration for graph learning and brings a new paradigm. Qubit is in multiple states simultaneously, thus quantum memory and quantum searching can retrieve a group of items at the same time~\cite{lvovsky2009optical}. As a result, accessing graph-structured data at a low cost has become a reality.
On the other hand, multiple quanta interact with each other by entanglement, so the relationship of quanta can be described as a graph. When QGL algorithms utilize quantum theory as physical knowledge (e.g., quantum random walk), graph-structured data is naturally suitable. In addition, QGL algorithms are the natural solution for graph-structured quantum data (e.g., quantum chemistry datasets).

Various algorithms are proposed to optimize the graph learning process at a software level, and the technologies of high-performance computation accelerate that at a hardware level. However, all these methodologies alleviate the symptom since the limitation of classical hardware is decided by the objective physical lows. Only the quantum mechanism can radically solve the efficiency problem of graph learning. The quantum theory shows its full advantages by combining with graph learning, rather than any other machine learning areas.

\subsection{Quantum is Omnipotent for Graph Learning}
Although quantum theory leads to a revolution in graph learning, the reliability of QGL is still uncertain. At the current stage of QML, algorithms are mainly designed to deal with classical data in Euclidean space. The input and output data in quantum circuits are quantum states in Hilbert space. Consequently, it is a necessary process to encode classical data in Hilbert space. Unfortunately, space transformation induces a distortion problem, which means the distance relationship between nodes distorts after encoding graphs. Specifically, connected nodes have a distinct representation, while unconnected nodes have a similar one. Several approaches have been proposed, but it is still unclear which is the best way to encode graph-structured data ~\cite{lloyd2020quantum,havlivcek2019supervised}. Furthermore, theory-oriented and knowledge-driven QGL algorithms lack migration capability~\cite{huang2022quantum}, and data-driven models cannot fully achieve quantum advantages with classical data under the shortage of quantum datasets~\cite{bravyi2018quantum,daley2022practical}.

\section{A Roadmap}
Quantum graph learning is now in its very beginning stage. There exists huge potential but meanwhile also faces many challenges. In this section, we will discuss the future directions of quantum graph learning.

\subsection{Graph-structured Data Encoding}
The distortion problem interferes with the learning process of QGL because of the loss in attribute and structure information after encoding graphs to Hilbert space. Conventional encoding approaches (e.g., amplitude encoding) apply to other QML models. While for QGL, trainable mapping is a better choice to preserve the characteristics of original graphs~\cite{ai2022decompositional}. Moreover, some real-world graphs are more complicated (e.g., heterogeneous graphs and dynamic graphs), which may lead to more severe problems in encoding. Heterogeneous graphs contain different node types and multiple relations that require specific representation in graph learning~\cite{dong2020heterogeneous,DBLP:conf/ijcai/WuW0021}. In QGL, encoding heterogeneous graphs is challenging, especially in the case of limited qubits. The integration of various information on quantum circuits is also difficult. The changing nodes and edges in dynamic graphs at each snapshot make it hard to encode serialized graph data~\cite{zheng2019addgraph,jia2020graphsleepnet}. However, the implicit correlation between the evolution of both quantum and dynamical graphs may lead to new algorithms.

\subsection{Large-Scale Graph Decomposition}
The scale of real-world graphs grows exponentially, which puts forward a higher requirement for QGL by employing a more significant number of qubits~\cite{BEDRU2020100247}. However, the increasing number of qubits definitely will cause more noise in NISQ quantum circuits~\cite {Preskill2018quantumcomputingin,cerezo2021cost}, and make the simulation on classical computers longer~\cite{RevModPhys.86.153}. Therefore, it is impractical to handle large-scale graphs for QGL. A trade-off solution for current QGL is graph decomposition which becomes significant to balance graph integrity and algorithm practicability.

\subsection{QGL Theory Investigation}
Graph learning gains a lot from quantum mechanics, but there are no general paradigms or guidelines for QGL algorithm designing. There still needs to be fundamental and comprehensive graph learning and quantum theory to reveal their explicit connection. Most previous QGL algorithms are partially based on either graph learning or quantum mechanism, which reflect the incomplete reciprocity of the two. In particular, a portion of current research mainly focuses on physical knowledge (e.g., quantum random walk), while another portion mostly concentrates on the learning process (e.g., quantum GNN). More efforts should be devoted to building a bridge between quantum theory and graph learning.

\subsection{Noise-Resilient Algorithms}
Hardware noise is inevitable for QML~\cite{torlai2020machine}, but most studies neglect noises in the analytical calculations and numerical simulations~\cite{biamonte2017quantum,cerezo2022challenges}. Furthermore, the noise problem is more severe for QGL. In graph learning, nodes and their representation are related to their neighbors. The noise causes an inaccurate representation of each node and further harms its neighbors' representation iteratively. To carry out QGL tasks on actual quantum circuits, noise-resilient algorithms and optimizers are indispensable.

\section{Conclusion}
This survey unprecedentedly provides an insightful introduction to QGL from the perspectives of its theoretical basis, implementation methods, and development prospects. Starting from the characteristics of graph data and the bottleneck of graph learning, we first discuss the mutualism of quantum theory and graph learning. Then, existing QGL methods are introduced in three categories, i.e., quantum computing on graphs, quantum graph representation, and quantum circuits for graph neural networks. Afterward, we point out the pitfalls and future directions of QGL to provide a reliable guideline for researchers. To the best of our knowledge, this paper is the first work to systematically review QGL and provide insights into its frontiers.

QGL is now emerging and has significant development potential for future industrial and scientific fields. The emergence of QGL inspires the improvement of graph learning. Particularly, traditional graph learning models are plagued by various inevitable limitations, such as lack of interpretability and limited ability to handle complex graph data. At the same time, QGL combines known physical principles (quantum theory) with graph learning models to address the shortcomings of existing methods to a great extent. Although there are some challenges facing QGL, i.e., it is hard to implement QGL on large-scale graphs, it will undoubtedly become a major research direction in the coming future.

\appendix

\section*{Acknowledgments}
The authors would like to thank Prof. Huan Liu (Arizona State University) for his valuable comments.

\bibliographystyle{named}
\bibliography{ijcai22}

\begin{thebibliography}{}

\bibitem[\protect\citeauthoryear{Ai \bgroup \em et al.\egroup
  }{2022}]{ai2022decompositional}
Xing Ai, Zhihong Zhang, Luzhe Sun, Junchi Yan, and Edwin Hancock.
\newblock Decompositional quantum graph neural network.
\newblock {\em arXiv preprint arXiv:2201.05158}, 2022.

\bibitem[\protect\citeauthoryear{Araujo \bgroup \em et al.\egroup
  }{2021}]{araujo2021divide}
Israel~F Araujo, Daniel~K Park, Francesco Petruccione, and Adenilton~J
  da~Silva.
\newblock A divide-and-conquer algorithm for quantum state preparation.
\newblock {\em Scientific reports}, 11(1):1--12, 2021.

\bibitem[\protect\citeauthoryear{Bai \bgroup \em et al.\egroup
  }{2021}]{bai2021learning}
Lu~Bai, Yuhang Jiao, Lixin Cui, Luca Rossi, Yue Wang, Philip Yu, and Edwin
  Hancock.
\newblock Learning graph convolutional networks based on quantum vertex
  information propagation.
\newblock {\em {TKDE}}, 2021.

\bibitem[\protect\citeauthoryear{Bedru \bgroup \em et al.\egroup
  }{2020}]{BEDRU2020100247}
Hayat~Dino Bedru, Shuo Yu, Xinru Xiao, Da~Zhang, Liangtian Wan, He~Guo, and
  Feng Xia.
\newblock Big networks: A survey.
\newblock {\em Computer Science Review}, 37:100247, 2020.

\bibitem[\protect\citeauthoryear{Beer \bgroup \em et al.\egroup
  }{2021}]{beer2021quantum}
Kerstin Beer, Megha Khosla, Julius K{\"o}hler, and Tobias~J Osborne.
\newblock Quantum machine learning of graph-structured data.
\newblock {\em arXiv preprint arXiv:2103.10837}, 2021.

\bibitem[\protect\citeauthoryear{Biamonte \bgroup \em et al.\egroup
  }{2017}]{biamonte2017quantum}
Jacob Biamonte, Peter Wittek, Nicola Pancotti, Patrick Rebentrost, Nathan
  Wiebe, and Seth Lloyd.
\newblock Quantum machine learning.
\newblock {\em Nature}, 549(7671):195--202, 2017.

\bibitem[\protect\citeauthoryear{Bravyi \bgroup \em et al.\egroup
  }{2018}]{bravyi2018quantum}
Sergey Bravyi, David Gosset, and Robert K{\"o}nig.
\newblock Quantum advantage with shallow circuits.
\newblock {\em Science}, 362(6412):308--311, 2018.

\bibitem[\protect\citeauthoryear{Cao \bgroup \em et al.\egroup
  }{2015}]{DBLP:conf/cikm/CaoLX15}
Shaosheng Cao, Wei Lu, and Qiongkai Xu.
\newblock Grarep: Learning graph representations with global structural
  information.
\newblock In {\em {CIMK}}, pages 891--900, 2015.

\bibitem[\protect\citeauthoryear{Cerezo \bgroup \em et al.\egroup
  }{2021}]{cerezo2021cost}
Marco Cerezo, Akira Sone, Tyler Volkoff, Lukasz Cincio, and Patrick~J Coles.
\newblock Cost function dependent barren plateaus in shallow parametrized
  quantum circuits.
\newblock {\em Nature Communications}, 12(1):1--12, 2021.

\bibitem[\protect\citeauthoryear{Cerezo \bgroup \em et al.\egroup
  }{2022}]{cerezo2022challenges}
M~Cerezo, Guillaume Verdon, Hsin-Yuan Huang, Lukasz Cincio, and Patrick~J
  Coles.
\newblock Challenges and opportunities in quantum machine learning.
\newblock {\em Nature Computational Science}, 2(9):567--576, 2022.

\bibitem[\protect\citeauthoryear{Chen \bgroup \em et al.\egroup
  }{2022}]{chen2022novel}
Yanhu Chen, Cen Wang, Hongxiang Guo, et~al.
\newblock Novel architecture of parameterized quantum circuit for graph
  convolutional network.
\newblock {\em arXiv preprint arXiv:2203.03251}, 2022.

\bibitem[\protect\citeauthoryear{Cong \bgroup \em et al.\egroup
  }{2019}]{cong2019quantum}
Iris Cong, Soonwon Choi, and Mikhail~D Lukin.
\newblock Quantum convolutional neural networks.
\newblock {\em Nature Physics}, 15(12):1273--1278, 2019.

\bibitem[\protect\citeauthoryear{Daley \bgroup \em et al.\egroup
  }{2022}]{daley2022practical}
Andrew~J Daley, Immanuel Bloch, Christian Kokail, Stuart Flannigan, Natalie
  Pearson, Matthias Troyer, and Peter Zoller.
\newblock Practical quantum advantage in quantum simulation.
\newblock {\em Nature}, 607(7920):667--676, 2022.

\bibitem[\protect\citeauthoryear{Dong \bgroup \em et al.\egroup
  }{2020}]{dong2020heterogeneous}
Yuxiao Dong, Ziniu Hu, Kuansan Wang, Yizhou Sun, and Jie Tang.
\newblock Heterogeneous network representation learning.
\newblock In {\em {IJCAI}}, volume~20, pages 4861--4867, 2020.

\bibitem[\protect\citeauthoryear{Dunjko and Briegel}{2018}]{dunjko2018machine}
Vedran Dunjko and Hans~J Briegel.
\newblock Machine learning \& artificial intelligence in the quantum domain: a
  review of recent progress.
\newblock {\em Reports on Progress in Physics}, 81(7):074001, 2018.

\bibitem[\protect\citeauthoryear{Georgescu \bgroup \em et al.\egroup
  }{2014}]{RevModPhys.86.153}
I.~M. Georgescu, S.~Ashhab, and Franco Nori.
\newblock Quantum simulation.
\newblock {\em Reviews of Modern Physics}, 86:153--185, 2014.

\bibitem[\protect\citeauthoryear{Grover}{1996}]{DBLP:conf/stoc/Grover96}
Lov~K. Grover.
\newblock A fast quantum mechanical algorithm for database search.
\newblock In {\em STOC}, pages 212--219. ACM, 1996.

\bibitem[\protect\citeauthoryear{Havl{\'\i}{\v{c}}ek \bgroup \em et al.\egroup
  }{2019}]{havlivcek2019supervised}
Vojt{\v{e}}ch Havl{\'\i}{\v{c}}ek, Antonio~D C{\'o}rcoles, Kristan Temme,
  Aram~W Harrow, Abhinav Kandala, Jerry~M Chow, and Jay~M Gambetta.
\newblock Supervised learning with quantum-enhanced feature spaces.
\newblock {\em Nature}, 567(7747):209--212, 2019.

\bibitem[\protect\citeauthoryear{Henry \bgroup \em et al.\egroup
  }{2021}]{henry2021quantum}
Louis-Paul Henry, Slimane Thabet, Constantin Dalyac, and Lo{\"\i}c Henriet.
\newblock Quantum evolution kernel: Machine learning on graphs with
  programmable arrays of qubits.
\newblock {\em Physical Review A}, 104(3):032416, 2021.

\bibitem[\protect\citeauthoryear{Huang \bgroup \em et al.\egroup
  }{2021}]{huang2021power}
Hsin-Yuan Huang, Michael Broughton, Masoud Mohseni, Ryan Babbush, Sergio Boixo,
  Hartmut Neven, and Jarrod~R McClean.
\newblock Power of data in quantum machine learning.
\newblock {\em Nature communications}, 12(1):1--9, 2021.

\bibitem[\protect\citeauthoryear{Huang \bgroup \em et al.\egroup
  }{2022}]{huang2022quantum}
Hsin-Yuan Huang, Michael Broughton, Jordan Cotler, Sitan Chen, Jerry Li, Masoud
  Mohseni, Hartmut Neven, Ryan Babbush, Richard Kueng, John Preskill, et~al.
\newblock Quantum advantage in learning from experiments.
\newblock {\em Science}, 376(6598):1182--1186, 2022.

\bibitem[\protect\citeauthoryear{Jia \bgroup \em et al.\egroup
  }{2020}]{jia2020graphsleepnet}
Ziyu Jia, Youfang Lin, Jing Wang, Ronghao Zhou, Xiaojun Ning, Yuanlai He, and
  Yaoshuai Zhao.
\newblock Graphsleepnet: Adaptive spatial-temporal graph convolutional networks
  for sleep stage classification.
\newblock In {\em IJCAI}, pages 1324--1330, 2020.

\bibitem[\protect\citeauthoryear{Karniadakis \bgroup \em et al.\egroup
  }{2021}]{karniadakis2021physics}
George~Em Karniadakis, Ioannis~G Kevrekidis, Lu~Lu, Paris Perdikaris, Sifan
  Wang, and Liu Yang.
\newblock Physics-informed machine learning.
\newblock {\em Nature Reviews Physics}, 3(6):422--440, 2021.

\bibitem[\protect\citeauthoryear{Kishi \bgroup \em et al.\egroup
  }{2022}]{kishi2022graph}
Kaito Kishi, Takahiko Satoh, Rudy Raymond, Naoki Yamamoto, and Yasubumi
  Sakakibara.
\newblock Graph kernels encoding features of all subgraphs by quantum
  superposition.
\newblock {\em IEEE Journal on Emerging and Selected Topics in Circuits and
  Systems}, 12(3):602--613, 2022.

\bibitem[\protect\citeauthoryear{Liu \bgroup \em et al.\egroup
  }{2021}]{liu2021rigorous}
Yunchao Liu, Srinivasan Arunachalam, and Kristan Temme.
\newblock A rigorous and robust quantum speed-up in supervised machine
  learning.
\newblock {\em Nature Physics}, 17(9):1013--1017, 2021.

\bibitem[\protect\citeauthoryear{Lloyd \bgroup \em et al.\egroup
  }{2020}]{lloyd2020quantum}
Seth Lloyd, Maria Schuld, Aroosa Ijaz, Josh Izaac, and Nathan Killoran.
\newblock Quantum embeddings for machine learning.
\newblock {\em arXiv preprint arXiv:2001.03622}, 2020.

\bibitem[\protect\citeauthoryear{Lucas}{2014}]{lucas2014ising}
Andrew Lucas.
\newblock Ising formulations of many np problems.
\newblock {\em Frontiers in Physics}, page~5, 2014.

\bibitem[\protect\citeauthoryear{Luo \bgroup \em et al.\egroup
  }{2019}]{PhysRevLett.123.070505}
Yi-Han Luo, Han-Sen Zhong, Manuel Erhard, Xi-Lin Wang, Li-Chao Peng, Mario
  Krenn, Xiao Jiang, Li~Li, Nai-Le Liu, Chao-Yang Lu, Anton Zeilinger, and
  Jian-Wei Pan.
\newblock Quantum teleportation in high dimensions.
\newblock {\em Physical Review Letters}, 123:070505, 2019.

\bibitem[\protect\citeauthoryear{Lvovsky \bgroup \em et al.\egroup
  }{2009}]{lvovsky2009optical}
Alexander~I Lvovsky, Barry~C Sanders, and Wolfgang Tittel.
\newblock Optical quantum memory.
\newblock {\em Nature photonics}, 3(12):706--714, 2009.

\bibitem[\protect\citeauthoryear{Mernyei \bgroup \em et al.\egroup
  }{2022}]{mernyei2022equivariant}
P{\'e}ter Mernyei, Konstantinos Meichanetzidis, and Ismail~Ilkan Ceylan.
\newblock Equivariant quantum graph circuits.
\newblock In {\em {ICML}}, pages 15401--15420. PMLR, 2022.

\bibitem[\protect\citeauthoryear{Miao \bgroup \em et al.\egroup
  }{2022}]{DBLP:conf/icml/MiaoLL22}
Siqi Miao, Mia Liu, and Pan Li.
\newblock Interpretable and generalizable graph learning via stochastic
  attention mechanism.
\newblock In {\em {ICML}}, volume 162, pages 15524--15543. PMLR, 2022.

\bibitem[\protect\citeauthoryear{Nielsen and Chuang}{2002}]{nielsen2002quantum}
Michael~A Nielsen and Isaac Chuang.
\newblock Quantum computation and quantum information, 2002.

\bibitem[\protect\citeauthoryear{Pelofske \bgroup \em et al.\egroup
  }{2021}]{pelofske2021decomposition}
Elijah Pelofske, Georg Hahn, and Hristo Djidjev.
\newblock Decomposition algorithms for solving np-hard problems on a quantum
  annealer.
\newblock {\em Journal of Signal Processing Systems}, 93(4):405--420, 2021.

\bibitem[\protect\citeauthoryear{Preskill}{2018}]{Preskill2018quantumcomputingin}
John Preskill.
\newblock Quantum {C}omputing in the {NISQ} era and beyond.
\newblock {\em {Quantum}}, 2:79, 2018.

\bibitem[\protect\citeauthoryear{Schuld and Killoran}{2019}]{schuld2019quantum}
Maria Schuld and Nathan Killoran.
\newblock Quantum machine learning in feature hilbert spaces.
\newblock {\em Physical review letters}, 122(4):040504, 2019.

\bibitem[\protect\citeauthoryear{Schuld \bgroup \em et al.\egroup
  }{2020}]{schuld2020measuring}
Maria Schuld, Kamil Br{\'a}dler, Robert Israel, Daiqin Su, and Brajesh Gupt.
\newblock Measuring the similarity of graphs with a gaussian boson sampler.
\newblock {\em Physical Review A}, 101(3):032314, 2020.

\bibitem[\protect\citeauthoryear{Shor}{1994}]{DBLP:conf/focs/Shor94}
Peter~W. Shor.
\newblock Algorithms for quantum computation: Discrete logarithms and
  factoring.
\newblock In {\em {FOCS}}, pages 124--134. IEEE, 1994.

\bibitem[\protect\citeauthoryear{Tabi \bgroup \em et al.\egroup
  }{2020}]{tabi2020quantum}
Zsolt Tabi, Kareem~H El-Safty, Zs{\'o}fia Kallus, P{\'e}ter H{\'a}ga, Tam{\'a}s
  Kozsik, Adam Glos, and Zolt{\'a}n Zimbor{\'a}s.
\newblock Quantum optimization for the graph coloring problem with
  space-efficient embedding.
\newblock In {\em {ICQCE}}, pages 56--62. IEEE, 2020.

\bibitem[\protect\citeauthoryear{Tang \bgroup \em et al.\egroup
  }{2022}]{tang2022quantum}
Yehui Tang, Junchi Yan, and Hancock Edwin.
\newblock From quantum graph computing to quantum graph learning: A survey.
\newblock {\em arXiv preprint arXiv:2202.09506}, 2022.

\bibitem[\protect\citeauthoryear{Torlai and Melko}{2020}]{torlai2020machine}
Giacomo Torlai and Roger~G Melko.
\newblock Machine-learning quantum states in the nisq era.
\newblock {\em Annual Review of Condensed Matter Physics}, 11:325--344, 2020.

\bibitem[\protect\citeauthoryear{T{\"u}ys{\"u}z \bgroup \em et al.\egroup
  }{2021}]{tuysuz2021hybrid}
Cenk T{\"u}ys{\"u}z, Carla Rieger, Kristiane Novotny, Bilge Demirk{\"o}z,
  Daniel Dobos, Karolos Potamianos, Sofia Vallecorsa, Jean-Roch Vlimant, and
  Richard Forster.
\newblock Hybrid quantum classical graph neural networks for particle track
  reconstruction.
\newblock {\em Quantum Machine Intelligence}, 3(2):1--20, 2021.

\bibitem[\protect\citeauthoryear{Verdon \bgroup \em et al.\egroup
  }{2019}]{verdon2019quantum}
Guillaume Verdon, Trevor McCourt, Enxhell Luzhnica, Vikash Singh, Stefan
  Leichenauer, and Jack Hidary.
\newblock Quantum graph neural networks.
\newblock {\em arXiv preprint arXiv:1909.12264}, 2019.

\bibitem[\protect\citeauthoryear{Wang \bgroup \em et al.\egroup
  }{2022}]{wang2022quantum}
Dawei Wang, Kedi Zheng, Fei Teng, and Qixin Chen.
\newblock Quantum annealing with integer slack variables for grid partitioning.
\newblock {\em IEEE Transactions on Power Systems}, 2022.

\bibitem[\protect\citeauthoryear{Wu \bgroup \em et al.\egroup
  }{2021}]{DBLP:conf/ijcai/WuW0021}
Chuhan Wu, Fangzhao Wu, Yongfeng Huang, and Xing Xie.
\newblock User-as-graph: User modeling with heterogeneous graph pooling for
  news recommendation.
\newblock In {\em {IJCAI}}, pages 1624--1630, 2021.

\bibitem[\protect\citeauthoryear{Xia \bgroup \em et al.\egroup
  }{2021}]{9416834}
Feng Xia, Ke~Sun, Shuo Yu, Abdul Aziz, Liangtian Wan, Shirui Pan, and Huan Liu.
\newblock Graph learning: A survey.
\newblock {\em IEEE Transactions on Artificial Intelligence}, 2(2):109--127,
  2021.

\bibitem[\protect\citeauthoryear{Xia \bgroup \em et al.\egroup
  }{2022}]{DBLP:journals/tnse/XiaYLLL22}
Feng Xia, Shuo Yu, Chengfei Liu, Jianxin Li, and Ivan Lee.
\newblock {CHIEF:} clustering with higher-order motifs in big networks.
\newblock {\em {TNSE}}, 9(3):990--1005, 2022.

\bibitem[\protect\citeauthoryear{Xu \bgroup \em et al.\egroup
  }{2020}]{DBLP:conf/jcdl/XuYSRLP020}
Jin Xu, Shuo Yu, Ke~Sun, Jing Ren, Ivan Lee, Shirui Pan, and Feng Xia.
\newblock Multivariate relations aggregation learning in social networks.
\newblock In {\em {JCDL}}, pages 77--86. Association for Computing Machinery,
  2020.

\bibitem[\protect\citeauthoryear{Yan \bgroup \em et al.\egroup
  }{2022}]{yan2022towards}
Ge~Yan, Yehui Tang, and Junchi Yan.
\newblock Towards a native quantum paradigm for graph representation learning:
  A sampling-based recurrent embedding approach.
\newblock In {\em {SIGKDD}}, pages 2160--2168, 2022.

\bibitem[\protect\citeauthoryear{Yu \bgroup \em et al.\egroup
  }{2019}]{DBLP:journals/access/YuXZXAT19}
Shuo Yu, Jin Xu, Chen Zhang, Feng Xia, Zafer Al{-}Makhadmeh, and Amr Tolba.
\newblock Motifs in big networks: Methods and applications.
\newblock {\em {IEEE} Access}, 7:183322--183338, 2019.

\bibitem[\protect\citeauthoryear{Yu \bgroup \em et al.\egroup
  }{2020}]{DBLP:journals/csr/YuFZBXX20}
Shuo Yu, Yufan Feng, Da~Zhang, Hayat~Dino Bedru, Bo~Xu, and Feng Xia.
\newblock Motif discovery in networks: {A} survey.
\newblock {\em Computer Science Review}, 37:100267, 2020.

\bibitem[\protect\citeauthoryear{Yu \bgroup \em et al.\egroup }{2021}]{9052480}
Shuo Yu, Feng Xia, Yuchen Sun, Tao Tang, Xiaoran Yan, and Ivan Lee.
\newblock Detecting outlier patterns with query-based artificially generated
  searching conditions.
\newblock {\em TCSS}, 8(1):134--147, 2021.

\bibitem[\protect\citeauthoryear{Yu \bgroup \em et al.\egroup
  }{2022a}]{10.1145/3487553.3524718}
Shuo Yu, Huafei Huang, Minh~N. Dao, and Feng Xia.
\newblock Graph augmentation learning.
\newblock In {\em Companion Proceedings of the Web Conference 2022}, page
  1063–1072. Association for Computing Machinery, 2022.

\bibitem[\protect\citeauthoryear{Yu \bgroup \em et al.\egroup
  }{2022b}]{yu2022graph}
Shuo Yu, Jing Ren, Shihao Li, Mehdi Naseriparsa, and Feng Xia.
\newblock Graph learning for fake review detection.
\newblock {\em Frontiers in Artificial Intelligence}, 5, 2022.

\bibitem[\protect\citeauthoryear{Zhang \bgroup \em et al.\egroup
  }{2019}]{zhang2019quantum}
Zhihong Zhang, Dongdong Chen, Jianjia Wang, Lu~Bai, and Edwin~R Hancock.
\newblock Quantum-based subgraph convolutional neural networks.
\newblock {\em Pattern Recognition}, 88:38--49, 2019.

\bibitem[\protect\citeauthoryear{Zheng \bgroup \em et al.\egroup
  }{2019}]{zheng2019addgraph}
Li~Zheng, Zhenpeng Li, Jian Li, Zhao Li, and Jun Gao.
\newblock Addgraph: Anomaly detection in dynamic graph using attention-based
  temporal gcn.
\newblock In {\em IJCAI}, volume~3, page~7, 2019.

\end{thebibliography}

\end{document}